\UseRawInputEncoding
\documentclass[prx,twocolumn,superscriptaddress,floatfix,nopacs,nofootinbib,longbibliography]{revtex4-2}

\usepackage{graphicx,amsfonts,amssymb,amsmath,hyperref,enumerate}

\usepackage{algorithm}
\usepackage{algpseudocode}
\usepackage{pgfplots}
\usepackage{pgf}
\usepackage{lmodern}
\usepackage{import}
\usepackage{xr}
\usepackage{enumitem}
\usepackage{soul}
\usepackage[normalem]{ulem}
\usepackage{multirow}
\UseRawInputEncoding
\newif\ifhyper
\hypertrue
\ifhyper
\hypersetup{
   citecolor = {green},
   colorlinks = {true}, 
   urlcolor = {blue} 
}
\fi

\newcommand{\beq}{\begin{equation}}
\newcommand{\eeq}{\end{equation}}
\newcommand{\beqa}{\begin{eqnarray}}
\newcommand{\eeqa}{\end{eqnarray}}

\newcommand{\comment}[1]{}

\def\Longarrow{\protect\@lra}
\def\@lra{\relbar\joinrel\relbar\joinrel\relbar\joinrel%
          \relbar\joinrel\rightarrow}

\pgfplotsset{compat=1.18}

\begin{document} 

\title{Boosting Defect Detection in Manufacturing \\ using Tensor Convolutional Neural Networks}

\author{Pablo Mart\'{i}n-Ramiro}
\author{Unai Sainz de la Maza}
\affiliation{Multiverse Computing, Parque Cientifico y Tecnol\'{o}gico de Gipuzkua, Paseo de Miram\'{o}n, 170 3$^{\,\circ}$ Planta, 20014 Donostia / San Sebasti\'{a}n, Spain}

\author{Sukhbinder Singh}
\affiliation{Multiverse Computing, Centre for Social Innovation, 192 Spadina Avenue Suite 509, Toronto, ON M5T 2C2 Canada}

\author{Rom\'{a}n Or\'{u}s}
\affiliation{Multiverse Computing, Parque Cientifico y Tecnol\'{o}gico de Gipuzkua, Paseo de Miram\'{o}n, 170 3$^{\,\circ}$ Planta, 20014 Donostia / San Sebasti\'{a}n, Spain}
\affiliation{Donostia International Physics Center, Paseo Manuel de Lardizabal 4, E-20018 San Sebasti\'an, Spain}
\affiliation{Ikerbasque Foundation for Science, Maria Diaz de Haro 3, E-48013 Bilbao, Spain}

\author{Samuel Mugel}
\affiliation{Multiverse Computing, Centre for Social Innovation, 192 Spadina Avenue Suite 509, Toronto, ON M5T 2C2 Canada}

\begin{abstract}
Defect detection is one of the most important yet challenging tasks in the quality control stage in the manufacturing sector.
In this work, we introduce a Tensor Convolutional Neural Network (T-CNN) and examine its performance on a real defect detection application in one of the components of the ultrasonic sensors produced at Robert Bosch's manufacturing plants. Our quantum-inspired T-CNN operates on a reduced model parameter space to substantially improve the training speed and performance of an equivalent CNN model without sacrificing accuracy. More specifically, we demonstrate how T-CNNs are able to reach the same performance as classical CNNs as measured by quality metrics, with up to fifteen times fewer parameters and $4\%$ to $19\%$ faster training times. Our results demonstrate that the T-CNN greatly outperforms the results of traditional human visual inspection, providing value in a current real application in manufacturing.
\end{abstract}

\maketitle

\section{Introduction}
\label{sec:intro}

One of the main challenges in manufacturing is distinguishing between high-quality and defective components as they are assembled together to create a complete product. This task becomes specially challenging in mass production, where a production line manufactures a massive number of products with complex structures. Traditionally, the quality control task is performed through visual inspection of images of the products by experienced human inspectors. This task requires analyzing hundreds or thousands of images per hour to distinguish the subtle features that usually characterize defective pieces. On top of human fatigue's impact on the results, these are usually subjective and hard to quantify. Therefore, it is crucial to automatize the quality control process beyond human inspection to increase the accuracy and performance of defect detection and reduce the number of misclassified products.

The big data revolution has enabled the development of new algorithms and techniques that can be extremely useful in applications with extensive data. Modern deep learning techniques have found a wide variety of applications in multiple fields, such as image classification \cite{krizhevsky2017imagenet}, object recognition \cite{wang2016deep}, or object detection \cite{8627998}. Convolutional neural networks (CNN) \cite{lecun1989} have been highly successful in image classification for their ability to extract and learn image's most relevant features (colors, shapes, and other patterns) from different classes. In particular, these techniques have been used with success in important applications in the manufacturing sector such as defect detection \cite{yang2019real, tabernik2020segmentation, yang2020, bhatt2021image}. However, defect detection in real-world environments is a challenging task. The reason for this is that defects are often subtle, diverse and sometimes hard to identify, which requires model architectures that are sufficiently complex to capture these small details and learn the features that characterize defects. This could be a potential issue because it is often related to models with a large number of parameters.
Despite their success in image vision tasks, CNNs are known to be over-parameterized, containing a significant number of parameters when working with large amounts of complex data \cite{du2018, soltanolkotabi2019}. This represents a bottleneck in the speed and the accuracy of the CNNs, increasing the computational resources required for training and inference times, which in turn, reduces the quality of results. There are multiple applications in sectors such as energy (smart grids \cite{chen2019, feng2021}), healthcare (patient monitoring \cite{rahman2021, hartmann2022}), communications (mobile networks \cite{abbas2017, mao2017}) and manufacturing (quality control \cite{dai2019industrial, kubiak2022possible}, predictive maintenance \cite{hafeez2021edge, nain2022towards}) that highly benefit from having models deployed as closed to the originating data source as possible, which is known as \emph{edge computing} \cite{satyanarayanan2017emergence}. In these applications, it is essential to develop models that show good performance and have a high computational efficiency and a small use of resources at the same time, so that they can be deployed on small edge computing or FPGA devices. For this reason, it is crucial to reduce the number of parameters in a CNN without sacrificing performance.

Reducing the number of parameters in a CNN is, in principle, a non-trivial task. Trimming the network by blindly discarding fractions of parameters regardless of the amount of information they learned would lead to a substantial loss of accuracy and misclassification of the images. One of the most extended approaches for reducing the number of network parameters in an efficient way is pruning \cite{li2016pruning, he2017channel}, which removes the weights or filters of the network that small values and thus have little contribution to the information learned by the network. A key ingredient for reducing the number of parameters in a CNN is to target the correct corner of the parameter space and only discard the parameters that have the least importance for learning. Quantum-inspired tensor network methods \cite{Orus2014, Verstraete2008} are good candidates to excel at this task, providing efficient and systematic ways of decomposing large tensors, i.e., ones used for describing most of the modern ML techniques. Tensor decompositions like canonical polyadic (CP) decomposition \cite{Hars1970}, or singular value decomposition (SVD) \cite{delathauwer2000} and its extension to higher-order tensors, i.e., Tucker decomposition \cite{Tucker1966}, are able to factorize multi-dimensional tensors, discarding parts of the original tensor that are less correlated and therefore can be discarded \cite{DeLathauwer2006}. This factorization methods have been used in a variety of areas like image vision \cite{tai2015convolutional, kim2015compression, novikov2015tensorizing, Panagakis2021}, component analysis \cite{Lu2014book}, dictionary learning \cite{Bahri2019}, and regression models \cite{Guo2012}. However, most of these applications are developed and tested on standard datasets and may not be completely representative of performance on real industrial environments. In this work, we use tensor factorization schemes and ideas from quantum-inspired tensor network methods \cite{Orus2014, Verstraete2008} to improve the efficiency of CNNs, reducing the number of parameters in the trainable weight tensors by keeping only the essential amount of network parameters for capturing the critical features and correlations in data. The value of the resulting Tensor Convolutional Neural Networks (T-CNN) is demonstrated on a real image-based defect detection application in the manufacturing sector.

To demonstrate the potential of our approach, we test the T-CNN on a real quality control application to detect defective components ultrasonic sensors in the production lines of Robert Bosch. Producing high-quality ultrasonic sensors is of crucial importance for their key roles in modern vehicles. Ultrasonic sensors are designed to perform fast and highly precise nearby obstacle detection, supporting drivers with tasks like maneuvering in narrow situations and parking, and enabling emergency braking functions at low speeds through faster reaction to obstacles \cite{borenstein1988obstacle}. These sensors are also used for collision avoidance in automated mobile machinery in the logistics, construction, and agriculture sectors. For their wide application in our daily life, manufacturing high-quality ultrasonic sensors is extremely important. To achieve this, we construct and apply a T-CNN to detect defects in one of the components of the manufactured ultrasonic sensors using an image dataset that contains thousands of examples collected from multiple production lines. Our results show that the T-CNN significantly outperforms traditional human-inspection at quality control. In addition, the T-CNN shows the same performance in terms of quality metrics as a regular CNN, but it provides a substantial advantage in terms of number of parameters and training time, which results in a more efficient use of computational resources.

The paper is organized as follows: In Section.~\ref{sec:problem}, we provide an overview of the defect detection problem that we tackle in the manufacturing process of ultrasonic sensors. In Section.~\ref{sec:methods}, we introduce the main concepts of tensor networks and present our methodology for building a T-CNN. In Section.~\ref{sec:experimental_setup}, we discuss the architectures of the CNN and T-CNN models, the training procedure and the quality metrics we used for assessing the accuracy and performance of the models. Results of our study are presented and discussed in Section.~\ref{sec:results}. Finally, Section.~\ref{sec:conclusion} is devoted to the discussion and conclusions.

\section{Problem overview}
\label{sec:problem}

In this section, we present an overview of the problem and a description of the dataset and the features that characterize defective pieces. The product component that is inspected in the quality control stage is a piezoelectric with two wires that are welded to it. The welding process can present multiple types of subtle defects, which should be caught at this quality control to prevent these defective components from reaching the next step of the manufacturing process. Therefore, the problem can be naturally formulated as a binary image classification problem, with the goal of distinguishing between high-quality pieces and defective pieces. In the following subsections, we will describe the dataset in more detail and formulate our approach to the problem.

\subsection{Dataset description}
\label{sec:dataset}

With the rise of big data and Industry 4.0 \cite{lasi2014industry}, manufacturing companies are collecting the vast amount of data that are continuously produced along the production process. The dataset we use in this application is a small fraction of this data, which has been manually labeled with the goal of training a supervised model to identify defective components. The dataset contains a total number of 11728 labeled images with a resolution of 1280x1024 pixels each. The produced components present nine possible types of defects, that may be labeled from 1 to 9 as: (1) broken piezoelectric, (2) weak weld, (3) strong weld, (4) misplaced weld, (5) piezoelectric in wrong position, (6) debris, (7) broken wire, (8) non-assessable image, and (9) short, long or no wire.

This problem presents multiple challenges. First, the data have been collected from multiple twin production lines, which adds an extra degree of complexity to the problem due to significant variations in illumination conditions and slightly different camera locations. This could be a potential issue for the model because the data distributions may differ significantly, making the learning process more difficult and introducing potential undesired biases to the model. Furthermore, the distribution of each type of defect and the absolute number of each of them differ in each production line. This makes it harder to train a unique model that classifies all the different defect types while maintaining good performance for all the production lines. Finally, piezoelectric suppliers may also vary from one line to another. This means defects can be different for each supplier, with different structures and material compounds, and therefore more challenging to identify. Taken together, all these features of the data can have a clear impact on performance, which will be overcome by introducing a data preprocessing strategy that will be explained in more detail in next subsection, and a data augmentation strategy during training that will be presented in next section.

\subsection{Problem Formulation}
\label{sec:problem_formulation}

Since the dataset has been constructed from a collection of images from multiple production lines, the distribution of each type of defect and the absolute number of each of them differ in each production line. This results in a lack of data for some defect classes, which we address by formulating the problem as a binary classification problem, grouping all types of failures into one single class.

Furthermore, the various illumination conditions in each production line are harmonized by introducing a data preprocessing stage that standardizes the color in all images, making them more homogeneous. More specifically, we increase the contrast of all images to make color more uniform and enhance shapes and borders in the piezoelectric component, which are key for distinguishing between high-quality and defective components. Furthermore, we resize all images to a resolution of $256$x$256$ and normalize the values of pixels.

\section{Tensor convolutional neural networks} 
\label{sec:methods}

In this section, we describe how to use quantum-inspired tensor network methods to improve the efficiency of CNN architectures. After introducing basic concepts tensor networks, we present a methodology for constructing a T-CNN and describe how the number of parameters is reduced compared to a CNN.

\subsection{Convolutional Neural Networks}

In this subsection, we present a brief overview of CNNs and explain the building blocks of their architectures and how convolutional layers are used for feature extraction. A classical CNN is composed of two main ingredients \cite{gu2018recent}. First, a single or multi-layer feature extraction network in which the most relevant features of an input image or data are extracted via a series of convolutional and pooling layers. Second, a classification (regression) network in which the learned features are processed by a sequence of fully connected layers to predict the label of the image or data. Each convolutional layer in the feature extraction network is responsible for learning a specific property of the image such as colors, edges, cracks, etc. that all together, define the features of a given image. On top of that, there are pooling layers that can be used to reduce the size of the learned representations by replacing blocks of data with their average or maximum values. The learned features are then fed to the classification network which uses those features to make a decision and predict the corresponding label of the image.

The number of parameters in the convolutional layer can still be very large even after pooling. On top of that, for complicated image structures, one may need to use many convolutional layers to capture different features. This will eventually over-parameterize the model and impose computational complexity on the training of the model, leading to a long training time, misclassification, and loss of accuracy. Therefore, it is important to find a good balance between the expressive power of the network and the number of parameters it contains. Ideally, the number of parameters should be reduced in a way that only parameters that contain key information are kept, while redundant parameters are discarded without influencing the performance and accuracy of the model. In the following subsections, we will describe how tensor network methods can be used to find compressed representations of the initial parameter space to achieve this goal.

\begin{figure}[t!]
\centerline{\includegraphics[width=8cm]{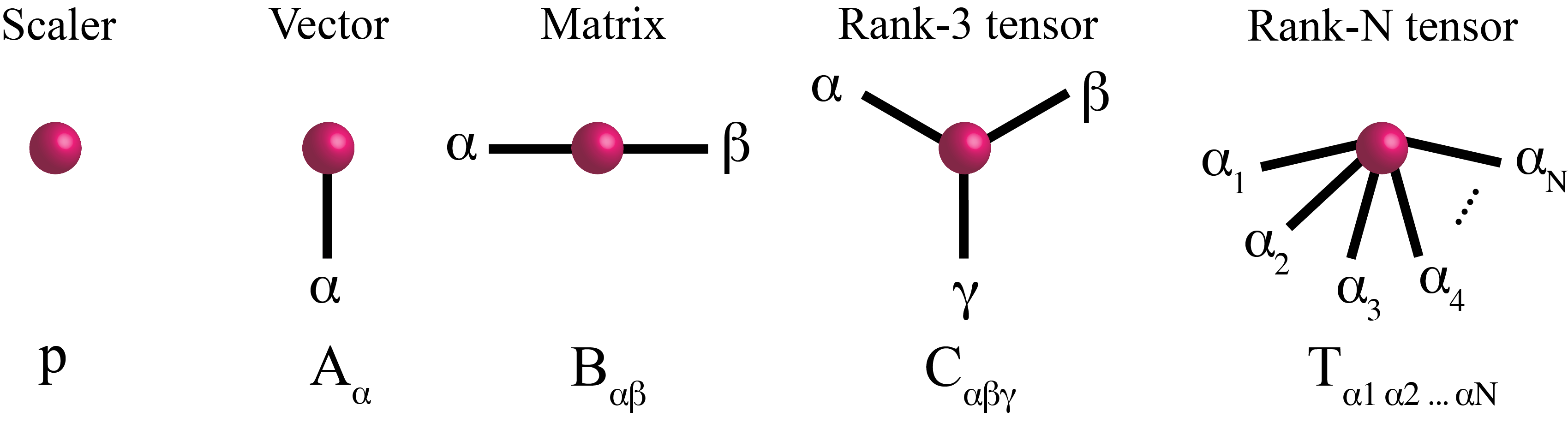}}
\caption{Tensor network diagrams. Each rank-$n$ tensor is denoted by an object with $n$ connected links each of which represents an individual dimension of the tensor. A scalar, a vector, a matrix and a rank-$n$ tensor have respectively, $0$, $1$, $2$, $n$ connected legs.}
\label{Fig:tn_diagrams}
\end{figure}

\subsection{Tensor network basics}

Tensor networks (TN) \cite{Orus2014, Verstraete2008} initially emerged in physics with the focus on providing efficient representation for the ground-state of quantum systems and a basis for some of the well-established numerical techniques such as density matrix renormalization group (DMRG) \cite{White1992, White1993} and time evolution block-decimation (TEBD) \cite{Vidal2003, Vidal2004}. However, due to its high potential for efficient data representation and compression, novel applications of TN are emerging in different branches of data science such as  machine learning (ML) and optimization \cite{tai2015convolutional, kim2015compression, novikov2015tensorizing, Panagakis2021, lebedev2014speeding, astrid2017cp, yu2017long, cao2018tensorizing}. In particular, it has been shown recently that TNs can be very successful in ML tasks such as classification \cite{Stoudenmire2016SupervisedLW}, clustering \cite{Stoudenmire_2018}, anomaly detection \cite{Wang2020} and even for solving partial differential equations with neural networks \cite{Patel2022}. Our goal in this section is to demonstrate how TNs can enhance and boost the efficiency, accuracy, and speed of CNNs by providing an efficient representation for the trainable weights of the convolutional layers. To this end, we first review the basic concepts of the TN and familiarize the reader with the concepts and notations that are widely used in the physics community for working with TNs.

A tensor is a multi-dimensional array of complex numbers represented by $T_{\alpha\beta\gamma\ldots}$ in which the subscripts denote different tensor dimensions and the number of these dimensions corresponds to the tensor rank. In a similar manner and in order to ease the burden of working with the mathematical notation of tensors, one can use tensor network diagrams \cite{Orus2014} to represent tensors. As illustrated in Figure~\ref{Fig:tn_diagrams}, a rank-$n$ tensor is an object with $n$ connected legs so that a scalar, a vector, and a matrix are objects with zero, one, and two connected legs, respectively. This diagrammatic notation is then generalized to rank-$n$ tensors with $n$ legs each corresponding to a tensor dimension.

\begin{figure}[t!]
\centerline{\includegraphics[width=8cm]{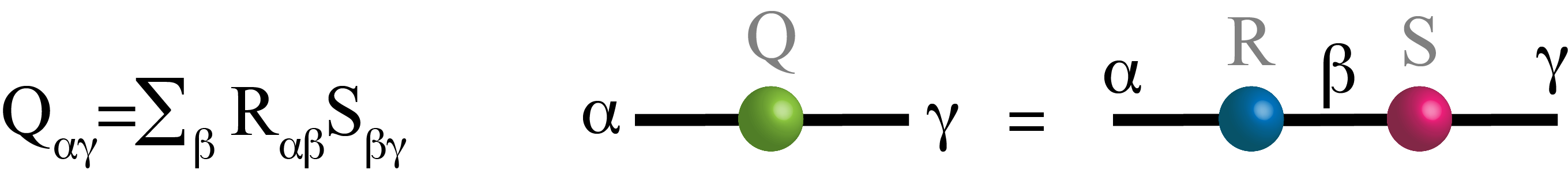}}
\caption{Contraction of tensors which is equivalent to tensor trace over their shared indices is represented graphically by connecting the shared links. Here, $R$ and $S$ tensors are connected along the shared leg $\beta$. This contraction operation is equivalent to matrix multiplication.}
\label{Fig:contraction}
\end{figure}

The TN diagrams not only represent the tensors but also represent tensor contractions, which is the generalization of matrix multiplication to rank-$n$ tensors. For example, contraction of two rank-$2$ tensors, i.e., matrices $R_{\alpha\beta}$ and $S_{\beta\gamma}$ along the dimension $\beta$ can be represented diagrammatically by connecting the two tensors along their shared $\beta$ legs, as depicted in Figure~\ref{Fig:contraction}. Equivalently, the contraction of these tensors can be represented mathematically as
\begin{equation}
Q_{\alpha\gamma} = {\rm tTR}(R_{\alpha\beta} S_{\beta\gamma}) = \sum_\beta R_{\alpha\beta} S_{\beta\gamma},    
\end{equation}
where the ${\rm tTR}$ is the tensor trace overs shared indices (tensor legs).

\begin{figure}[t!]
\centerline{\includegraphics[width=8cm]{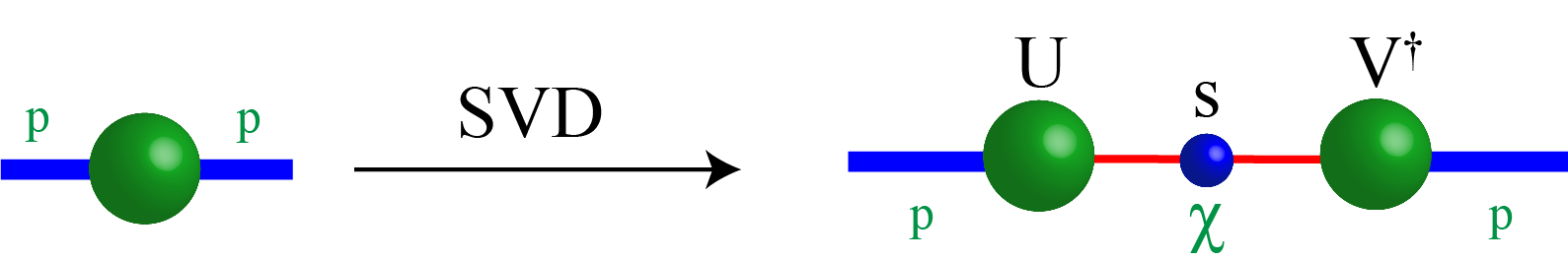}}
\caption{The SVD decomposition of a matrix to two rank-$3$ tensors which are called matrix product operator (MPO).}
\label{Fig:svd}
\end{figure}

Another remarkable feature of TNs that can be inspiring for designing many TN algorithms is tensor decomposition, among which the singular value decomposition (SVD) is of great importance. SVD is a matrix decomposition operation that factorizes the input matrix as the product of two unitary matrices and a diagonal matrix of singular values. Figure~\ref{Fig:svd} demonstrates an example of an SVD which factorizes a $p\times p$ matrix into the $p\times \chi$ unitary matrix $U$, the $\chi\times \chi$ diagonal matrix $S$ of singular values, and the $\chi\times p$ unitary matrix $V^{\dagger}$ such that
\begin{equation}
M=U S V^\dagger = L R, \quad \quad L = U \sqrt{S}, \quad \quad R = \sqrt{S} V^\dagger
\end{equation}

The size of new interconnecting red link $\chi$, depicted in red, controls the amount of classical correlation (or entanglement in the quantum case) between the left and right pieces. Considering the example of Figure~\ref{Fig:svd}, in such a decomposition there exist at most $\chi_{\rm max} = p$ singular values in the diagonal matrix $S$. The number of non-zero elements in the $S$ matrix correspond to the amount of entanglement (correlation) between the $L$ and $R$ tensors. If the data are weakly correlated usually, most of the singular values are close to zero and can be discarded so that only the $\chi\le\chi_{\rm max}$ largest singular values are kept. This truncation scheme provides a clever and accurate way of data compression by only discarding the most irrelevant information which plays no significant role in the correlation among factorized pieces. This implies that the interconnecting tensor dimensions which emerge in the decomposition are the relevant degrees of freedom for capturing the correlations between parameters of the tensor network.

The concept of singular value decomposition can be extended to the higher-order tensors by using the high-order SVD (HOSVD) \cite{DeLathauwer2006}. The HOSVD, which is also called Tucker decomposition \cite{Tucker1966} in mathematics community, factorizes a rank-$n$ tensor to the product of $n$ factor matrices and a rank-$n$ core tensor as illustrated in Figure~\ref{Fig:tucker} for a rank-$4$ tensor. In the next subsection, we will use the Tucker decomposition to tensorize a CNN.

\subsection{Tensorizing a CNN}\label{sec:tensorizing}

\begin{figure}[t!]
\centerline{\includegraphics[width=8cm]{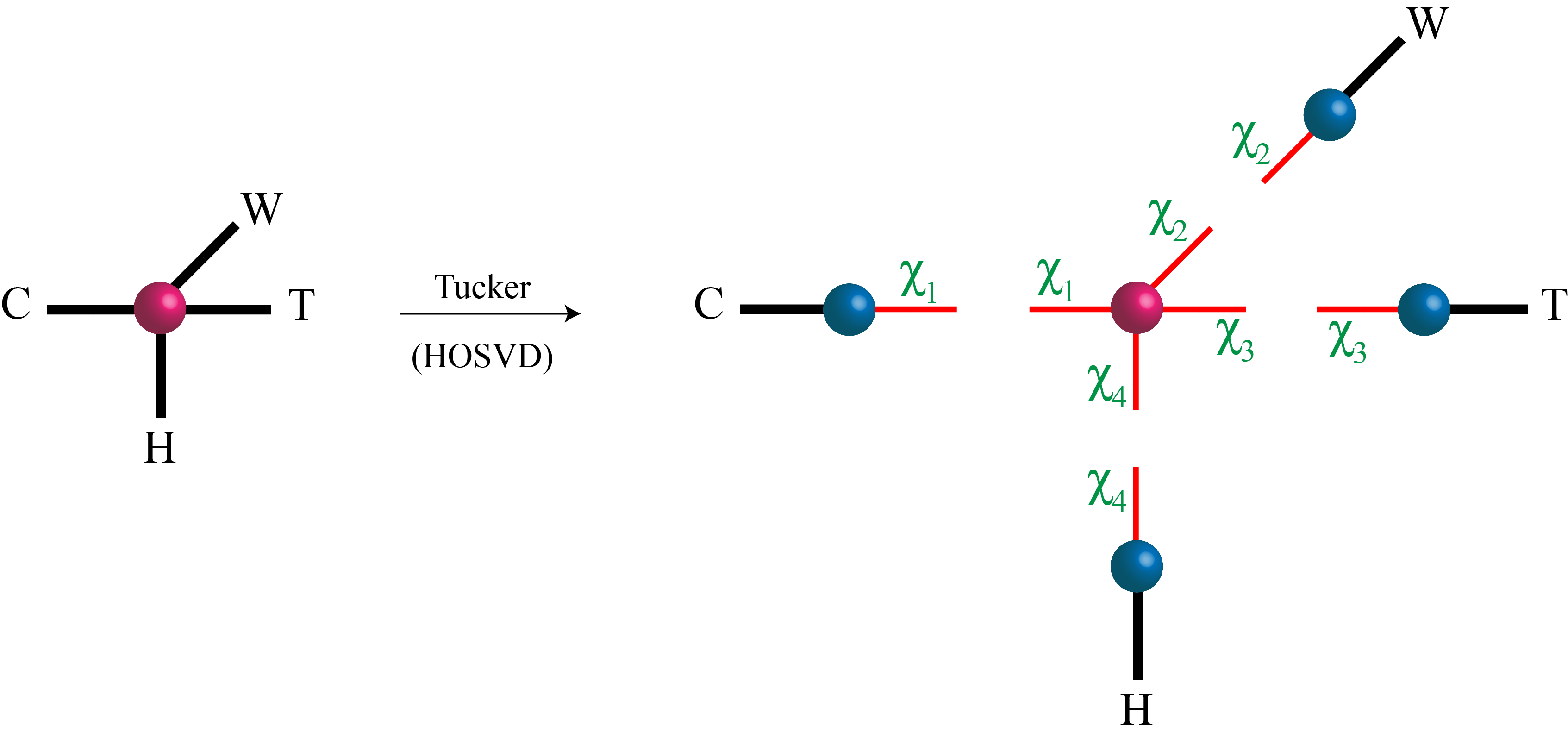}}
\caption{The HOSVD (Tucker) factorization of the convolution weight tensor to a core tensor and four factor matrices. The $\chi_i$s are the factorization (truncation) ranks of the factor matrices.}
\label{Fig:tucker}
\end{figure}

Let us now see how we can leverage the tensor network and Tucker decomposition to tensorize a CNN. Let us consider a 2D CNN. Each convolutional layer in a classic CNN contains a rank-4 weight tensor as depicted in Figure~\ref{Fig:tucker}. The four dimensions of the convolution tensor, i.e., $C$, $T$, $W$, $H$ correspond to the number of input channels (filters), output channels, width, and height of the features in that layer, respectively. The training process of the CNN amounts for finding the optimum parameters for the weight tensors in each layer. Depending on the complexity and size of the problem, convolutional weight tensors can be both numerous and large, implying a huge number of trainable parameters. Storing parameters in memory and fine-tuning and training over such a large parameter space can in principle be computationally very expensive and, at some point, beyond the reach of many devices such as mobile phones or electronic instruments with small memory and battery. It is therefore of great importance to reduce the number of parameters in a clever way without sacrificing accuracy. 

The parameters of the convolution weight tensors of a CNN can be shrunk by factorizing the weight tensors, keeping the most relevant information learned by the network while discarding the irrelevant parts. The factorized tensors are obtained from tensor decomposition of the original weight tensor by applying Tucker decomposition. In Tucker decomposition, the original tensor is approximated by the contraction of a core tensor and four other factor matrices, each of which has a truncated interconnecting dimension, as illustrated in Figure~\ref{Fig:tucker}. The size of truncated dimensions, which are also called factorization ranks, control the data compression rate and the size of the reduced parameter space after factorization. The CNN model with factorized convolutional layer is then called Tensor CNN or T-CNN. 

Training a T-CNN is in practice performed with automatic differentiation and backpropagation during gradient descent optimization. However, the main difference between the training of a classic CNN and a T-CNN is that for the former, a single big rank-$4$ weight tensor is trained for each convolutional layer, whereas the latter involves updating four smaller factorized matrices and a core tensor within each layer. Once updated, the factorized matrices and the core tensors of each convolutional layer are contracted and the information is fed to the next layer. Let us further stress that the size of factorized tensors in each convolutional layer is a new hyper-parameter of the T-CNN and it is fixed once we tensorize the model. While, depending on the optimization technique, one in principle has the possibility of adaptive changing the size of factorized tensors during training \cite{Jahromi2022}, here we kept the size of factorized tensors fixed.

The training process of the T-CNN can be done following two different approaches. The first approach consists in training a CNN and tensorize the pre-trained model using Tucker decomposition. This results in a model with one core tensor and four orthogonal factorized matrices\footnote{Note that the core tensor and the four factorized matrices obtained after Tucker decomposition are not unique.}. The four factorized matrices can be interpreted as the principal components in each dimension, whereas the core tensor would describe interactions between the different dimensions \cite{wall2003singular}. In this approach, key correlations in data are kept by taking the principal singular values of the Tucker decomposition while irrelevant parameters are truncated, resulting in a new tensor representation that defines a compressed parameter space where only essential information is retained. The second approach consists in training the model from scratch on the compressed parameter space defined by the new tensor representation. In this approach, the trained model parameters are not the result of a Tucker decomposition directly, but found by optimizing the model in the new tensor representation compressed space directly. Note that the global minimum of the model loss function is the same in both cases, but it is reached differently. In this work, we implement the T-CNN following the second approach, which has the advantage of accessing the compressed parameter space directly while having faster training times.

\subsection{Parameter Counting}

In a classical CNN, the model parameters belong in principle, to three categories: i) parameters in convolutional layers, $N_c$; ii) bias parameters, $N_b$; and iii) parameters of the classification (regression) layer, $N_r$; which sums up to 
\begin{equation}
\label{eq:N_CNN}
    N_{\rm CNN} = N_c + N_b + N_r.
\end{equation}
More specifically, the number of trainable parameters in the weight tensor of a convolutional layer (see Figure~\ref{Fig:tucker}-left) is $N_c= C\times W\times T\times H$. As pointed out in the previous subsection, in order to tensorize a CNN and reduce the number of parameters, we factorize the convolution weight tensors. To have an estimate of the number of parameters of the factorized tensors, we introduce four factorization ranks for the output dimension of the factor matrices which are upper-bounded by the size of that dimension, i.e.,
\begin{equation}
\chi_1 \le C, \quad \quad \chi_2 \le W, \quad \quad \chi_3 \le T, \quad \quad \chi_4 \le H.     
\end{equation}
The total number of parameters in the factorized convolutional layers of the T-CNN is therefore given as the sum of parameters of the four factor matrices and the parameters of the core tensors, which in total reads:
\begin{equation}
    N_c^f = C\times\chi_1+W\times\chi_2+T\times\chi_3+H\times\chi_4 + (\chi_1\times\chi_2\times\chi_3\times\chi_4).
\end{equation}
Adding $N_c^f$ to the parameters of bias and the classification layer, the total number of parameters in a T-CNN is given by:
\begin{equation}
\label{eq:N_TCNN}
    N_{\text{T-CNN}} = N_c^f + N_b + N_r.
\end{equation}

We can further define a metric that indicates the compression ratio in the parameters of the T-CNN as compared to that of its classical CNN counterpart as:
\begin{equation}
\label{eq:R_TCNN}
    C_r = \frac{N_c}{N_c^f}.
\end{equation}
In a practical model for image classification, generally $N_c^f\ll N_c$, giving rise to a huge reduction factor in the number of model parameters.

\section{Experimental setup} 
\label{sec:experimental_setup}

In this section, we describe the experimental setup we used to build, train and test the model in the defect detection application approached in this work.

\subsection{Model architecture}

The architecture of the reference CNN model is a simplified version of the standard VGG16 architecture \cite{simonyan2014very}. The structure of the CNN model has been optimized for best performance in this particular problem by incrementally adding and tuning layers until quality metrics stop improving as measured on validation data. Once the reference CNN model is optimized, we tensorize this architecture by replacing regular convolutional layers by tensor convolutional layers and train the new T-CNN model from scratch, as described in previous section. Importantly, the rank configurations of each layer act as a hyperparameter that must be optimized for best performance. In this work, we considered various round numbers for the rank configurations to show how the T-CNN performs as measured by quality metrics, compression ratio (defined in Eq.~\ref{eq:R_TCNN}) and training time in different scenarios, each of them defined by the number of parameters of the T-CNN model. This experimental setup allows us to establish a direct comparison between the new T-CNN and a CNN, and analyze the capabilities of a non-optimized T-CNN as measured by different metrics. We remark that both the architecture and the rank configurations in all layers of the T-CNN may be tuned for optimal performance.

To build a T-CNN, different tensor factorization schemes such as MPS, Tucker or CP decomposition can be used to factorize the weight tensors of convolutional layers. However, we found that Tucker decomposition performs best with a more stable training and better results. Let us further point out that in addition to the convolutional layer, the classification layer of the CNN can also be tensorized using standard tensorization schemes for fully connected dense layers. For example, this layer may be replaced by a tensor neural network layer or tensor regression layers. In particular, tensor regression layers are interesting for their direct connection to the convolutional layers, without the need to flatten the output of convolutional layers. We observed that introducing tensor regression layers did not have a positive impact on performance metrics in this case and thus these layers were not included in the final model.

\subsection{Training setup}

All models have been implemented in PyTorch \cite{paszke2017automatic}, using techniques such as mixed-precision training to reduce memory usage and computational requirements. All models were trained for 80 epochs using the Adam \cite{kingma2014adam} optimizer. The learning rate was varied from $3 \cdot 10^{-4}$ to $1 \cdot 10^{-6}$ using a Multi-Step fixed learning rate scheduler. Models were trained on a NVIDIA T4 GPU, a graphic card equipped with 16GB of memory, and specially designed for hardware acceleration for deep learning training and inference.

We have divided the entire dataset in three splits, where the $80\%$ is used for training, $10\%$ for validation, and $10\%$ for testing. In this way, out of the total number of 11728 labeled images, 9382 are used for training, 1173 are used for validation and 1173 for testing. The first split is only used for training the models, the second one is used for model selection and  hyperparameter tuning, and the last one is only used for testing model performance in unseen data and generate all results.

Current deep neural networks models like ours require a lot of data to obtain good results and prevent overfitting. For this reason, we implemented a set of augmentations when training the model to increase the diversity of data, creating new training examples by slightly modifying the existing ones. This process eliminates spurious correlations that are specific to each production line, such as any remaining color patterns and the slightly different camera orientations, making the resulting model more robust. The details of this process were presented in Section~\ref{sec:experimental_setup}. In particular, all images where augmented by using a random combination of color, resized crop and cutout augmentations, each with a predefined probability that was fine tuned for best performance. The preprocessing and augmentation procedures make the model more robust to specific features of each production line. However, models may still be able to learn any remaining spurious correlations that characterize each production line. Even if this was the case, we explicitly checked that training on all available data yields better performance in all lines individually.

Furthermore, the number of images of non-defective pieces in the original dataset is higher than that of defective ones. This class imbalance may encourage the model to only learn to correctly identify the majority class, i.e., non-defective images. To alleviate this issue, we collected more defective images and used a weighted random sampling technique where we give more importance to the minority class, which is more efficient than other common procedures such as oversampling.

\subsection{Performance metrics}

As we have an imbalanced dataset, performance metrics like accuracy can be misleading. For this reason, we measure model performance using precision, recall and F1 score as \emph{quality metrics}, defined as:
\begin{equation}
\begin{aligned}
    &\text{Precision} = \frac{\text{TP}}{\text{TP} + \text{FP}} \, , \\
    &\text{Recall} = \frac{\text{TP}}{\text{TP} + \text{FN}} \, , \\
    &\text{F1} = \frac{2 \cdot (\text{Precision} \cdot \text{Recall})}{\text{Precision} + \text{Recall}} \, .
\end{aligned}
\end{equation}

Furthermore, we also introduce the \emph{slip-through} metric \cite{Damm2006FaultsslipthroughA}, which measures the fraction of defective images escaping detection in relation to the total number of defective images at this stage of the quality control process. The slip-through is thus defined as:
\begin{equation}
    \text{slip-through} = 1 - \text{Recall} \, .
\label{eq:slip}
\end{equation}
This metric is especially relevant for measuring the effectiveness of the quality control process in real mass production manufacturing environments.

One of the main goals of improving CNNs with Tensor Networks is to make them more efficient and discard all the unnecessary information, i.e., parameters. In practice, there are several ways to measure the efficiency of the neural networks. We choose a combination of two metrics, one that measures the parameter reduction made to the original CNN, i.e., the compression ratio defined in Eq.~\ref{eq:R_TCNN}, and another metric to compare the training times of CNN and T-CNN with different ranks. For the latter, we define the training time improvement T as the training time advantage of the T-CNN divided by the total training time of the CNN, which reads as
\begin{equation}
    \text{T} = \frac{T_{\text{CNN}} - T_{\text{T-CNN}}}{T_{\text{CNN}}} \times 100 \, ,
\end{equation}
where $T_{\text{CNN}}$ and $T_{\text{T-CNN}}$ are the mean training times of several CNN and T-CNN models.

\section{Results} 
\label{sec:results}

This section presents a detailed analysis of results on the performance of the T-CNN for classifying high-quality and defective piezoelectric components.

\subsection{Performance of the T-CNN }

In this subsection, we analyze how T-CNNs with different rank configurations in tensor convolutional layers compare to their CNN counterpart in terms of quality metrics, number of parameters and training time.

As described in Section~\ref{sec:tensorizing}, each tensor convolutional layer is parameterized by a 4-rank tensor. There are two dimensions, $r_{in}$ and $r_{out}$, that can be related to the number of input and output channels of the layer, whereas $h$ and $w$ can be related to the size of the convolutional filter.
We develop models with various rank configurations and evaluate their performance in terms of three different types of metrics: quality metrics (precision, recall and F1 score), compression ratio and training time improvement in relation to a CNN.
In order to enable direct comparison, the architecture of the T-CNN is the same as that of the optimized CNN, but regular convolutional layers are replaced by tensor convolutional layers.
The optimized CNN contains convolutional layers with various numbers of filters. By contrast, all tensor convolutional layers in the T-CNN have the same fixed rank configuration, which acts as an upper bound for the rank of the tensor that defines the layer. This choice is motivated by the fact that the first layers of the CNN network have a smaller number of filters, whereas the last layers have a larger number of filters. This results in lower compression for the information learned by the first layers of the T-CNN and larger compression in the final layers, where the number of learned feature representations is higher and their structures may be more complex. However, note that the rank configuration in all tensor convolutional layers should be optimized for best performance.

\begin{table*}[t!]
\centering
\begin{tabular}{c|c|c|c|c|c|c|c|c|c}
\hline
Model & $r_{in}$ & $r_{out}$ & $h$ & $w$ & Precision & Recall & F$1$ & Compression & \begin{tabular}[c]{@{}c@{}}Training time\\ improvement\end{tabular} \\
\hline
CNN & \multicolumn{4}{c|}{-} & $0.972 \pm 0.005$ & $0.933 \pm 0.006$ & $0.952 \pm 0.005$ & x$1$ & x$1$ \\
\hline
\multirow{5}{*}{T-CNN} & $96$ & $96$ & $3$ & $3$ & $0.968 \pm 0.007$ & $0.935 \pm 0.006$ & $0.951 \pm 0.004$ & x$1.2$ & $4\%$ \\
& $64$ & $64$ & $3$ & $3$ & $0.968 \pm 0.008$ & $0.932 \pm 0.007$ & $0.950 \pm 0.004$ & x$1.9$ & $10\%$ \\
& $32$ & $32$ & $3$ & $3$ & $0.962 \pm 0.010$ & $0.930 \pm 0.007$ & $0.946 \pm 0.006$ & x$4.6$ & $16\%$ \\
& $16$ & $16$ & $3$ & $3$ & $0.948 \pm 0.011$ & $0.931 \pm 0.008$ & $0.939 \pm 0.007$ & x$9.4$ & $19\%$ \\
& $8$  & $8$  & $3$ & $3$ & $0.928 \pm 0.012$ & $0.923 \pm 0.007$ & $0.925 \pm 0.008$ & x$15.7$ & $19\%$ \\
\hline
\end{tabular}
\caption{Performance of T-CNN models with multiple rank configurations as measured by quality metrics, compression ratio and training time improvement in relation to the optimized CNN. Columns two to five correspond to the maximum ranks for the four dimensions of the underlying tensor in each tensor convolutional layer of the T-CNN. For each model, results are presented as the mean over twenty different random seeds and therefore twenty different network initializations, and uncertainties correspond to one standard deviation.}
\label{tab:tcnn_vs_cnn}
\end{table*}

A detailed analysis of results for the T-CNN is presented in Table~\ref{tab:tcnn_vs_cnn}.
We consider five different T-CNN models, each defined by a maximum rank configuration in each tensor convolutional layer: $(96, 96, 3, 3)$, $(64, 64, 3, 3)$, $(32, 32, 3, 3)$, $(16, 16, 3, 3)$ and $(8, 8, 3, 3)$. These round values for the rank configurations of the T-CNN were selected to present a broad picture of the balance between quality metrics, compression ratio and training times that can be achieved by the model.
Our analysis provides two key results. First, a T-CNN with rank $(96, 96, 3, 3)$ shows the same performance as the optimized CNN in terms of quality metrics, slightly favoring recall over precision, while having $20\%$ less parameters and similar training time. Moreover, T-CNNs with ranks $(64, 64, 3, 3)$ and $(32, 32, 3, 3)$ are still able to achieve the same quality metrics, while adding significant x$1.9$ and x$4.6$ reductions in the number of model parameters, with $10\%$ and $16\%$ faster training times, respectively. This result suggests that the compressed parameter space defined by tensor convolutional layers is better equipped to capture the key correlations in data, keeping the essential amount of information and encoding it more effectively, while ignoring the irrelevant noise, than the CNN.

Second, as the maximum rank for the tensor convolutional layers is decreased, there exists a trade-off between model performance as measured by the quality metrics and computational efficiency (compression factor and training time improvement).
This compromise is expected, since networks with larger ranks are expected to learn more than the essential amount of information, whereas networks with large compression factors do not have the capacity to capture some important patterns, which yields slightly reduced quality metrics.
Importantly, our results show that there exists a set of optimal rank configurations that bring together the best of the two worlds, with the T-CNN achieving identical performance as the optimized CNN in quality metrics, while having a much smaller number of parameters and faster training times.
It is important to note that a T-CNN with a small rank configuration of $(8, 8, 3, 3)$ for each convolutional layer achieves large compression ratios with only a negligible drop in performance, demonstrating that the tensor convolutional layer is able to capture the essential amount of information, ignoring unimportant correlations in data. Note that this performance is expected to drop when removing layers in the model. We remark that we analyzed rank configurations of size $(96, 96, 3, 3)$, $(64, 64, 3, 3)$, $(32, 32, 3, 3)$, $(16, 16, 3, 3)$ and $(8, 8, 3, 3)$. These round numbers are used to test the T-CNN over multiple ranks and they should be tuned for optimal performance. A systematic hyperparameter optimization that finds the best rank configuration for all layers in the network is encouraged to find the best possible model in each application.

The advantages offered by the T-CNN could be key in multiple industrial and scientific applications. 
In applications where achieving the highest possible quality metrics is essential, a T-CNN may be able to deliver the same results as its classical counterpart, while using up to five times less parameters and training $4\%$ to $16\%$ times faster. This is often the case in applications such as quality control systems in manufacturing, image-based disease detection in healthcare, or obstacle detection in autonomous driving vehicles. 
By contrast, in applications where computational resources are limited or models have to be deployed in small edge computing or FPGA devices, a T-CNN may be built using nine to sixteen times less parameters and trained $20\%$ faster, while sacrificing $1 - 3\%$ in quality metrics as measured by the F1 score. This is often the case when models have to be deployed in small edge computing or FPGA devices, where data processing is done closer to where is needed with fast response times. While the specific values of quality metrics, compression ratios and training times depend on the specific application, our results suggest the T-CNN advantage shown in this work could extrapolate to other datasets. Furthermore, inference time is another important factor in manufacturing applications where processing a large number of samples in a short period of time is of utmost importance. This is especially important for small edge computing devices, where the computational power limits the maximal throughput of the model. In this case, we observe that the T-CNN model with rank configuration $(32, 32, 3, 3)$ has an average inference time of $143.4 \, \text{ms}$ for a batch of 128 images, which is similar to the $143.7 \, \text{ms}$ for the CNN model.

\begin{table*}[t!]
\centering
\begin{tabular}{c|cc|c|c|c|c|c|c|c}
\hline
& \multicolumn{2}{c|}{\begin{tabular}[c]{@{}c@{}}Confusion\\ matrix\end{tabular}} & Precision & Recall & F1 & AUC & Slip-through ($\%$) & Compression & \begin{tabular}[c]{@{}c@{}}Training time\\ improvement\end{tabular} \\
\hline
\multirow{4}{*}{CNN}
    & \multicolumn{1}{c|}{TN} & 768 & \multirow{4}{*}{0.964} & \multirow{4}{*}{0.951} & \multirow{4}{*}{0.958} & \multirow{4}{*}{0.984} & \multirow{4}{*}{$4.9\%$} & \multirow{4}{*}{x1}     & \multirow{4}{*}{x1} \\ \cline{2-3}
    & \multicolumn{1}{c|}{FP} & 14                     &                         &                        &                         &                        &                         &                   \\ \cline{2-3}
    & \multicolumn{1}{c|}{FN} & 19 
    &                         &                 
    &                         &                     
    &                         &                   \\ \cline{2-3}
    & \multicolumn{1}{c|}{TP} & 372                  
    &                         &                 
    &                         &                     
    &                         &                   \\ 
\hline

\multirow{4}{*}{\begin{tabular}[c]{@{}c@{}}T-CNN\end{tabular}}
    & \multicolumn{1}{c|}{TN} & 764 & \multirow{4}{*}{0.954} & \multirow{4}{*}{0.954} & \multirow{4}{*}{0.954} & \multirow{4}{*}{0.984} &\multirow{4}{*}{$4.6\%$} & \multirow{4}{*}{x4.6}  & \multirow{4}{*}{$16\%$} \\ \cline{2-3}
    & \multicolumn{1}{c|}{FP} & 18                     &                         &                        &                         &                        &                         &                   \\ \cline{2-3}
    & \multicolumn{1}{c|}{FN} & 18 
    &                         &                 
    &                         &                     
    &                         &                   \\ \cline{2-3}
    & \multicolumn{1}{c|}{TP} & 373                  
    &                         &                 
    &                         &                     
    &                         &                   \\ 
\hline
\end{tabular}
\caption{Performance of the best T-CNN model on test data for a rank configuration $(32, 32, 3, 3)$ as measured by quality metrics, compression ratio and training time improvement. In this case, we also include the AUC metric as a measure of model performance regardless of the decision threshold. Results for the best CNN are shown for comparison. In both cases, the selected models showed best performance as measured by the F1 score on validation data. All metrics are calculated at a fixed threshold of $0.2$, which is key for lowering the slip-through. For reference, the estimated slip-through of human inspection in a typical production line shift is $10\%$.}
\label{tab:best_tcnn_vs_cnn}
\end{table*}

In this defect detection application, it is essential to reach the highest possible quality metrics while maintaining a reasonable number of parameters and training times. With this in mind, motivated by the results presented in Table~\ref{tab:tcnn_vs_cnn} we identify the T-CNN model with rank configuration $(32, 32, 3, 3)$ as the optimal choice in this case, and present detailed results for the best T-CNN model on test data in Table~\ref{tab:best_tcnn_vs_cnn}. The best T-CNN model is the one showing best performance on validation data as measured by the F1 score. We now include results for the slip-through, a quality metric that we introduced in Eq.~\ref{eq:slip} and it is of most interest in the real production environment considered in this work. Our results show that the T-CNN and the optimized CNN are able to reach excellent performance as measured by quality metrics, with the T-CNN using almost five times less parameters and training $16\%$ times faster.

Importantly, the ability of the T-CNN model for classifying high-quality and defective pieces is substantially beyond human inspection, reducing the fraction of defective images escaping detection from $10\%$ to $4.6\%$. In the context of mass manufacturing, this improvement has significant economic and quality assurance benefits, as it leads to significant cost savings and enhanced product quality and reliability. Furthermore, integrating a T-CNN model in quality control systems can also lead to efficiency gains, as it would free up human operators from repetitive and time-consuming visual inspection tasks. This would allow companies to invest human resources on the areas of production where workers are essential: tasks that require high cognitive and creative abilities, problem-solving skills, and human expertise for decision making, leading to increased efficiency and productivity.



\subsection{Error analysis}
\label{sec:error_analysis}

This section contains a systematic error analysis of the results presented for the T-CNN in Table~\ref{tab:best_tcnn_vs_cnn}, with the goal of extracting valuable insights that can be used to improve the model in the future. For this purpose, we identified all defective images that are misclassified as false negatives and thus increase the slip-through. Then we analyzed these images in detail to determine the type of defect, to recognize any potential mislabeled images and to understand the reasons why each image has been misclassified.

Three conclusions can be extracted from this analysis on the eighteen images that have been labeled as defective and are misclassified by the T-CNN. First, one of them is clearly defective and could be classified correctly by lowering the decision threshold from $0.2$ to $0.1$, at the expense of a slightly larger amount of non-defective images being misclassified as defective (false positives). Second, there are another four images in the test set that have been mislabeled as defective. Therefore, annotating the samples is mandatory to improve model performance in this case. Furthermore, note that mislabeled images will also be present in the train and validation sets. This may introduce an undesired bias to the model that makes the whole training process harder, as the model is trying to learn from some contradictory information. However, this is usually the case with standard datasets \cite{ekambaram2017finding, northcutt2021pervasive}.

Finally, there is a total of four pieces with defects of type weak, strong or displaced weld, and another five pieces with defects of type debris. In both cases, these classes of defects seem to be harder to detect. For the former, this can be explained by the subtleties that characterize this subset of defects. For the latter, the distinctive feature of this kind of defect is a small dark area on top of the piezoelectric component. Although debris is easier to detect to the human eye, the preprocessing and color augmentations could be distorting this defect. Performance on these types of defects could be improved by adding more samples to the training dataset. From the remaining four images, two of them have a broken piezoelectric weld, whereas the other two have a broken wire. In this case, we cannot explain why the model fails to classify these obvious defects correctly.

As discussed in last section, defect types that are underrepresented in the training set may have larger error ratios. Therefore, adding more defective images from these misrepresented classes to the training stage is crucial to improve overall model performance. Furthermore, curating the entire dataset would likely result in better performance.

\section{Conclusion and outlook}
\label{sec:conclusion}

Quality control automation and detection of imperfect and defective products in mass production is one the crucial yet challenging tasks of most applications in the manufacturing sector. Traditional approaches based on human visual inspection of product images are prone to error, mainly because defects are often subtle and hard to identify. Therefore, this has a negative impact both on the quality of results and on human inspectors, with human fatigue being an issue due to the repetitive nature of the task. In this work, we integrated tensor network methods into the structure of a CNN to build a Tensor Convolutional Neural Network (T-CNN). We showed how the T-CNN model can be used to automate and improve defect detection in a real quality control process of Robert Bosch 's manufacturing plants.

The T-CNN model is built by replacing the convolutional layers of a CNN by factorized tensor convolutional layers layers based on high-order tensor decomposition. We showed that for a CNN, the rank-$4$ weight tensors of the convolutional layers can be factorized into four factor matrices and one core tensor, in which interconnecting dimensions of the factorized tensors control the data compression ratio and the amount of correlation between the factorized pieces. We trained the T-CNN model from scratch on the compressed parameter space defined by the new tensor representation.
Our results show that the T-CNN has two advantages compared to an optimized CNN: a significantly smaller number of parameters and faster training time, while having comparable learning capacity as measured by quality metrics. These advantages provide significant benefits in terms of computational cost, training time, robustness, and interpretability.

Our results show how T-CNNs with high ranks can achieve the same performance as a CNN as measured by precision, recall and F1 score metrics, while offering moderate improvements in terms of the number of parameters and training time. For lower ranks, the T-CNN is still able to keep the same high quality metrics as a CNN, while offering significant improvements in the number of parameters and training speed. More specifically, we showed how a T-CNN with maximum ranks of $(32, 32, 3, 3)$ in each layer attains comparable quality metrics, with $4.6$ times fewer parameters and $16 \%$ faster training times.
This results suggest that the compressed parameter space defined by the tensor convolutional layers is better equipped to capture the key correlations in data, keeping the essential amount of information and encoding it more effectively, while ignoring the irrelevant noise, compared to a CNN.

Furthermore, our model is capable of substantially outperforming human inspection, reducing the fraction of defective images escaping detection from $10\%$ to $4.6\%$. In a typical production line, this $54\%$ improvement leads to significant cost savings and enhanced product quality and reliability. Moreover, integrating a T-CNN model in quality control systems would allow companies to free up human resources for tasks where humans are essential; such as those that require creativity, problem-solving skills, and human expertise for decision making, leading to increased efficiency and productivity.
While the specific performance in terms of quality metrics, compression ratios and training times depend on the specific application, our results suggest the advantage of the T-CNN shown in this work could extrapolate to other datasets. In addition, a key feature of the model is that once the model is trained, classifying new images is extremely fast compared to human inspection. In particular, a batch of 128 images is classified in $143.4 \, \text{ms}$.

Importantly, the factorization rank of the T-CNN plays the role of a hyperparameter, which controls the number of trainable parameters in the tensor convolutional layers. In this work, we chose to present a broad analysis and consider multiple round rank configurations. However, a systematic hyperparameter optimization should be performed to find the best rank configuration for all layers in the T-CNN for optimal performance.

Finally, the T-CNN has excellent potential for integrating into a production environment for fast real-time defect detection in quality control processes. The T-CNN model is fast, accurate and efficient in terms of memory and resource usage, enabling deployment on small devices such as mobile phones, edge computing devices and even FPGAs. Furthermore, the smaller number of parameters of the T-CNN makes training more energy efficient and, therefore, more suited for devices with small energy resources and batteries.

\section*{Acknowledgments}
\label{sec:acknowledgments}
\vspace{-5pt}
We would like to acknowledge and thank the technical teams and field engineers both at Robert Bosch and Multiverse Computing for helpful discussions.

%

\end{document}